\documentclass[11pt, a4paper, logo, copyright, nonumbering]{wechat}
\usepackage[numbers, square, sort&compress]{natbib}
\usepackage{dblfloatfix}
\usepackage{ulem}
\usepackage{caption}
\definecolor{citecolor}{HTML}{1976D2}
\hypersetup{
    colorlinks=true,
    linkcolor=black,
    filecolor=magenta,
    urlcolor=blue!50!black,
    citecolor=citecolor,
}
\usepackage{dramatist}
\usepackage{xspace}
\usepackage{pifont}
\usepackage{multirow}
\usepackage{tcolorbox}
\usepackage{xltabular}
\usepackage{longtable}
\usepackage{hyperref}
\usepackage{diagbox}
\usepackage{makecell}
\usepackage{amsmath}   
\usepackage{amssymb}   
\usepackage{amsfonts}  
\usepackage{bm}
\usepackage{lineno}

\usepackage[bottom]{footmisc}

\usepackage{CJKutf8}
\usepackage{subfigure}
\usepackage{setspace}
\usepackage{subcaption} 

\usepackage{titlesec}

\usepackage{tcolorbox}
\usepackage{fontawesome5}
\usepackage{enumitem}

\usepackage{booktabs}
\usepackage{multirow}
\usepackage[table,xcdraw]{xcolor}
\usepackage{graphicx}

\usepackage{algorithm}
\usepackage{algorithmic}
\usepackage{subcaption}

\tcbset{
    contributionbox/.style={
        colback=blue!3,
        colframe=blue!50!black,
        boxrule=0.5pt,
        arc=3pt,
        left=6pt,
        right=6pt,
        top=6pt,
        bottom=6pt,
        fonttitle=\bfseries
    }
}

\titlespacing*{\paragraph}
  {0pt}                   
  {0.5ex plus 1ex minus .2ex} 
  {1em}            

\makeatletter
\def\@BTrule[#1]{%
  \ifx\longtable\undefined
    \let\@BTswitch\@BTnormal
  \else\ifx\hline\LT@hline
    \nobreak
    \let\@BTswitch\@BLTrule
  \else
     \let\@BTswitch\@BTnormal
  \fi\fi
  \global\@thisrulewidth=#1\relax
  \ifnum\@thisruleclass=\tw@\vskip\@aboverulesep\else
  \ifnum\@lastruleclass=\z@\vskip\@aboverulesep\else
  \ifnum\@lastruleclass=\@ne\vskip\doublerulesep\fi\fi\fi
  \@BTswitch}
\makeatother

\addto\extrasenglish{
}

 {\begin{list}{}%
         {\setlength{\leftmargin}{#1}}%
         \item[]%
 }
 {\end{list}}

\reportnumber{001} 

\title{\centering FlashPrefill: Instantaneous Pattern Discovery and Thresholding for Ultra-Fast Long-Context Prefilling}

\author{
    \vspace{0.5em}
    {\large \bfseries
    Qihang Fan$^{1,2,3,\ast}$, Huaibo Huang$^{1, 2,\dagger}$, Zhiying Wu$^{3}$, Juqiu Wang$^{2,3,\ast}$, 
    } \\ \vspace{0.6pt}
    {\large \bfseries
    Bingning Wang$^{3,\ddagger}$, Ran He$^{1, 2}$
    } \\ \vspace{1.0em}
    
    {\normalsize \normalfont
    $^{1}$MAIS\&NLPR, CASIA \quad
    $^{2}$UCAS \quad
    $^{3}$WeChat, Tencent 
    }
}

\renewcommand{\phi}{\varphi}

\renewcommand{\geq}{\geqslant}

\renewcommand{\epsilon}{\varepsilon}
\renewcommand{\imath}{\mathrm{i}}

\newlength{\restsubwidth}
\newlength{\restsubheight}
\newlength{\restsubmoreheight}
\newcommand{\rest}[2]{%
        \settowidth{\restsubwidth}{\ensuremath{#2}}
        \settoheight{\restsubheight}{\ensuremath{{}_{#2}}}
        \ensuremath{{#1\hskip 0.5pt}_{\vrule\kern2pt\parbox[b][%
        4pt][b]{\the\restsubwidth}{%
                        \ensuremath{{}_{#2}}}}}
        }

\github{https://github.com/qhfan/FlashPrefill}

\begin{abstract}
Long-context modeling is a pivotal capability for Large Language Models, yet the quadratic complexity of attention remains a critical bottleneck, particularly during the compute-intensive prefilling phase. While various sparse attention mechanisms have been explored, they typically suffer from either significant search latency or insufficient sparsity. In this paper, we propose FlashPrefill, a framework enabling ultra-fast prefilling via instantaneous pattern discovery and thresholding. FlashPrefill leverages a fast block-searching technique to simultaneously locate dynamic vertical, slash, and block-sparse attention patterns. Crucially, it introduces a dynamic thresholding mechanism that bypasses the prohibitive overhead of sorting or accumulating attention scores while effectively eliminating the long-tail distribution to enhance sparsity. Extensive evaluations demonstrate that FlashPrefill achieves a substantial leap in efficiency, delivering an unprecedented $\mathbf{27.78\times}$ speedup on 256K sequences. Notably, unlike existing methods that incur efficiency degradation on shorter contexts, FlashPrefill maintains a $\mathbf{1.71\times}$ speedup even at a 4K context length, demonstrating its robustness and practical utility across varying sequence scales.
\end{abstract}

\begin{document}
\begin{CJK*}{UTF8}{gbsn}

\maketitle

\enlargethispage{1cm}

\newcommand\blfootnote[1]{%
  \begingroup
  \renewcommand\thefootnote{}\footnote{#1}%
  \addtocounter{footnote}{-1}%
  \endgroup
}

\blfootnote{$^\dagger$ Corresponding Author.}
\blfootnote{$^\ddagger$ Project Leader.}
\blfootnote{$^\ast$ Work done during internship at WeChat.}

\section{Introduction}

Recent years have witnessed the rapid evolution of Large Language Models (LLMs), with the emergence of numerous highly capable models serving a massive user base across a diverse range of tasks~\cite{llama,llama2,llama3}. However, due to the inherent quadratic complexity of self-attention—the foundational operator of the Transformer architecture—LLMs often incur prohibitive time overhead when processing long-context sequences~\cite{attention}. This bottleneck is particularly salient during the compute-intensive prefill stage.

\begin{figure*}[t]
    \centering
    \begin{minipage}[b]{0.48\linewidth}
        \centering
        \includegraphics[width=\linewidth]{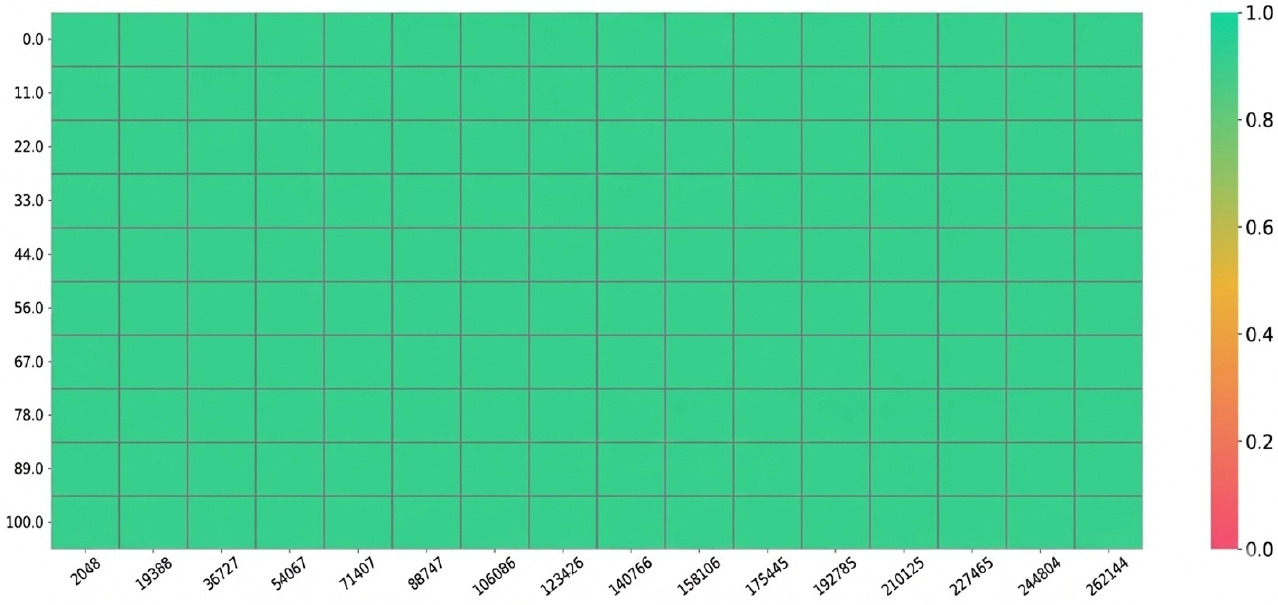}
        \caption{"Needle In A Haystack" evaluation of Qwen3-30B-A3B-Instruct-2507 using FlashPrefill across context lengths ranging from 2K to 256K.}
        \label{fig:niah_qwen3}
    \end{minipage}
    \hfill
    \begin{minipage}[b]{0.48\linewidth}
        \centering
        \includegraphics[width=\linewidth]{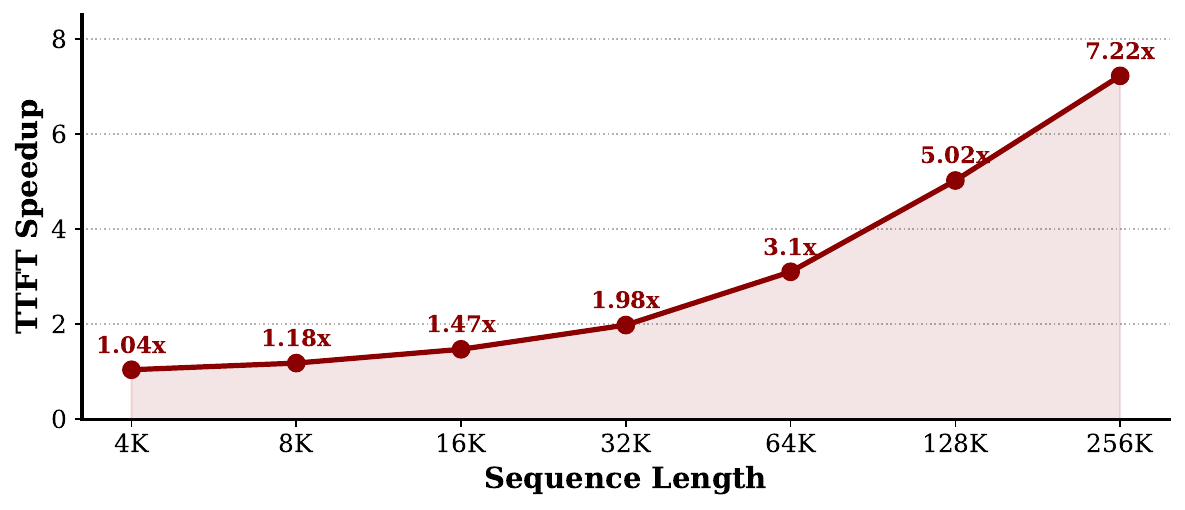}
        \caption{End-to-end Time-to-First-Token (TTFT) speedup relative to full attention on Qwen3-30B-A3B-Instruct-2507 within the vLLM framework.}
        \label{fig:e2e_speedup}
    \end{minipage}
\end{figure*}

To mitigate these challenges, various sparse attention mechanisms have been proposed, most of which adhere to a consistent underlying principle~\cite{minference,flexprefill,xattention}. These methods typically begin with a coarse-grained estimation of attention scores to identify salient blocks or structural patterns—such as vertical or diagonal (slash-like) stripes—using selection strategies like Top-$k$ or Top-$p$. By subsequently restricting fine-grained attention computation to these critical tokens, they effectively accelerate long-context prefilling while preserving maximum information fidelity.

Nevertheless, existing methodologies leave a substantial margin for efficiency improvements. First, the preliminary coarse-grained estimation of attention scores often introduces non-negligible computational latency. Second, selection strategies such as Top-$k$ or Top-$p$ necessitate explicit sorting of attention scores—either globally or locally—incurring high overhead on modern GPU architectures. Top-$p$ schemes even require a cumulative summation over the score distribution, an inherently sequential process that is resistant to parallelization and leads to significant end-to-end delays. Lastly, Top-$k$ and Top-$p$ heuristics struggle to effectively prune the long-tail distribution of tokens with marginal influence, resulting in incomplete sparsity and persistent computational redundancy.

In light of these considerations, we propose FlashPrefill, an ultra-fast prefill acceleration framework tailored for long-context scenarios. FlashPrefill features an Instantaneous Pattern Discovery stage that accurately identifies prevalent attention structures, such as vertical, slash, and block-wise patterns. To overcome the high latency typically associated with the discovery phase, we introduce a block-approximation strategy specifically designed to optimize the computation kernel. By streamlining memory access and parallelizing the discovery process through this block-level approximation, FlashPrefill reduces the discovery overhead to a negligible level, enabling the model to perceive the global attention landscape with extreme efficiency.

\begin{figure*}[ht]
    \centering
    \includegraphics[width=0.95\linewidth]{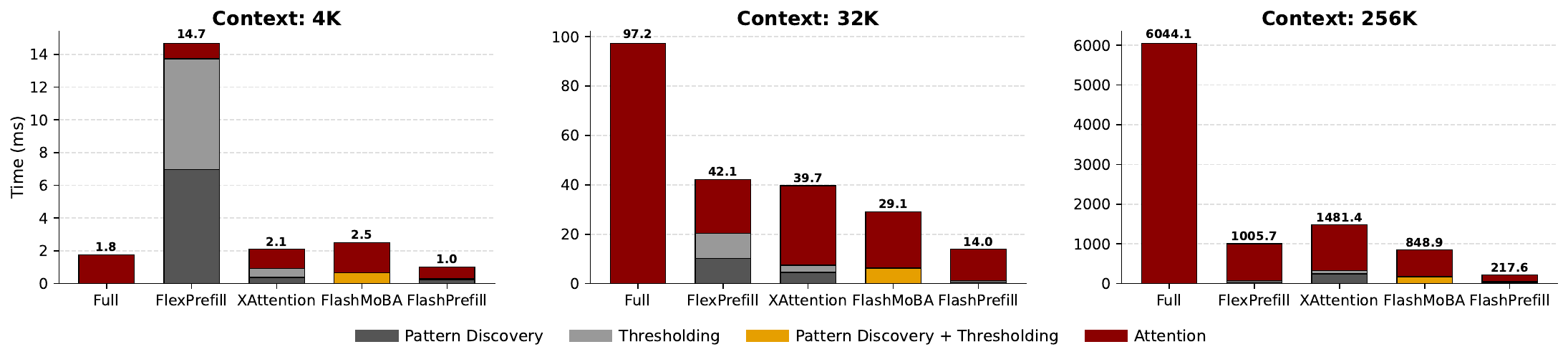}
    \caption{Execution time of different parts across different approaches. All results are measured on Qwen3-30B-A3B-Instruct-2507.}
    \label{fig:dis_thr}
\end{figure*}

Furthermore, FlashPrefill departs from traditional Top-$k$ or Top-$p$ block selection methods, which are often bottlenecked by significant sorting latencies and computational complexity. Instead, it utilizes a more efficient Max-based Dynamic Thresholding mechanism. This approach not only streamlines the identification of salient blocks but also adaptively addresses the "incomplete sparsity" problem typically caused by the heavy-tailed distribution of attention scores. By effectively filtering out redundant blocks without the need for exhaustive sorting, FlashPrefill ensures a more thorough sparse representation, thereby achieving substantial prefill acceleration while maintaining robust model performance across extensive context windows. Fig.~\ref{fig:dis_thr} illustrates the time breakdown of different components across various methods. FlashPrefill significantly reduces the time spent on each component.

\begin{figure*}[t]
    \centering
    \includegraphics[width=0.98\linewidth]{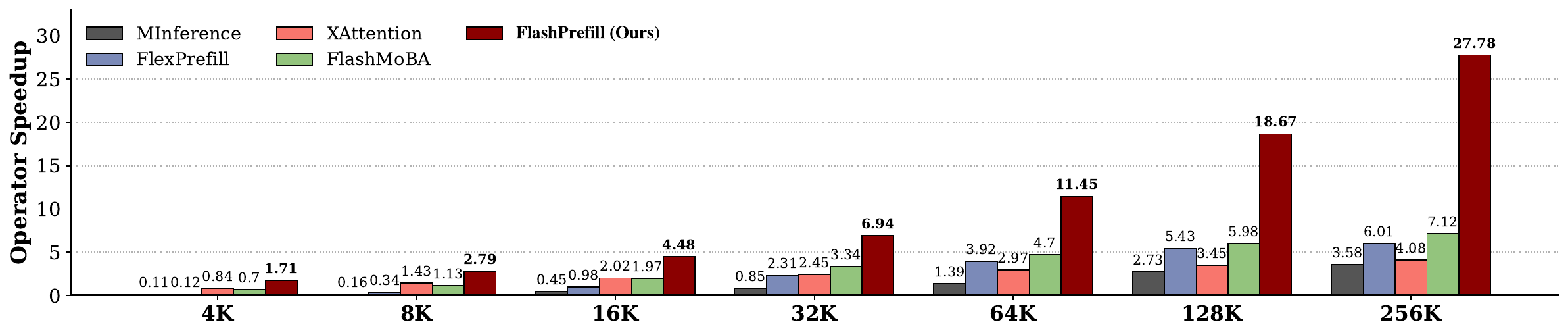}
    \caption{Comparative speedup of various operators relative to Flash Attention~\cite{flashattention2} on Qwen3-30B-A3B-Instruct-2507. FlashPrefill exhibits a dominant advantage, particularly in long-context scenarios.}
    \label{fig:attention_speedup}
\end{figure*}

We evaluated FlashPrefill across a diverse range of Large Language Models (LLMs) and Vision Language Models (VLMs), the results of which comprehensively underscore its superior performance. Taking Qwen3-30B-A3B-Instruct-2507 (which supports a maximum context length of 256K) as a representative example, Fig.~\ref{fig:attention_speedup} illustrates the speedup achieved by various operators across different sequence lengths. FlashPrefill demonstrates a significant advantage; even at a relatively short sequence length of 4K, it still delivers a $\mathbf{1.71\times}$ speedup. As the sequence length scales to 256K, FlashPrefill achieves a remarkable acceleration of $\mathbf{27.78\times}$.Furthermore, we integrated FlashPrefill into the vLLM inference framework to measure the end-to-end Time to First Token (TTFT). As shown in Fig.~\ref{fig:e2e_speedup}, FlashPrefill achieves a maximum end-to-end speedup of $\mathbf{7.22\times}$. Crucially, the experimental results on the "Needle In A Haystack" test for Qwen3, depicted in Fig.~\ref{fig:niah_qwen3}, indicate that FlashPrefill maintains nearly identical model performance with negligible loss in accuracy.

Our contributions can be summarized as follows:
\begin{itemize}
    \item We propose an Instantaneous Pattern Discovery method and introduce a block approximation strategy to accelerate kernel computation and reduce memory access overhead.
    \item We propose a Max-based Dynamic Thresholding method, which effectively eliminates the time overhead associated with local sorting in Top-$k$ or cumulative summation in Top-$p$ strategies. Simultaneously, it mitigates the impact of long-tail distributions on sparse attention.
    \item Leveraging the aforementioned strategies, we introduce FlashPrefill, an efficient approach for accelerating the long-context prefill stage. Its efficacy is demonstrated through rigorous evaluations across a diverse set of models and benchmarks.
\end{itemize}
\section{Related Works}

\paragraph{Large Language Models.}In recent years, the emergence of a wide array of Large Language Models (LLMs) has catalyzed rapid advancements across both academia and industry~\cite{qwen25technicalreport,qwen3technicalreport,qwentechnicalreport, yiopenfoundationmodels,qwen2.5-1m,glm2024chatglm,minimax01scalingfoundationmodels}. Furthermore, the rich prior knowledge inherent in LLMs can be leveraged to enhance Vision-Language Models (VLMs), facilitating the construction of intelligent multimodal agents~\cite{qwen25vltechnicalreport,qwen3vltechnicalreport,qwen2vlenhancingvisionlanguagemodels,fan2025sec,liu2023llava}. Beyond the classical Transformer architectures based on full attention, a growing number of novel architectures have emerged in modern LLMs to mitigate the prohibitive computational overhead incurred by full attention when processing long contexts. Such as the models use the sparse attention and linear attention~\cite{mamba,mamba2,yang2024deltanet,yang2024gdn,zhang2025kda,yang2024gla,sun2023retentivenetworksuccessortransformer,fan2024breaking,fan2024rect,mobamixtureblockattention,optimizingmixtureblockattention}.

\paragraph{Sparse Attention.}As a pivotal component of the Transformer architecture, the attention mechanism plays a crucial role in capturing dependencies~\cite{attention}. However, its quadratic complexity leads to significant computational overhead, especially in long-context scenarios. To address this, numerous sparse attention mechanisms have been proposed to reduce the computational cost of the attention layer~\cite{minference,flexprefill,xattention,mobamixtureblockattention,optimizingmixtureblockattention,native-sparse-attention,zhao2025infllmv2}. While a portion of these approaches necessitates explicit model training or fine-tuning~\cite{native-sparse-attention,zhao2025infllmv2}, others serve as training-free methods that can be seamlessly integrated into existing LLMs~\cite{flexprefill,minference,xattention}.
\begin{figure}[t]
    \centering
    \includegraphics[width=0.7\linewidth]{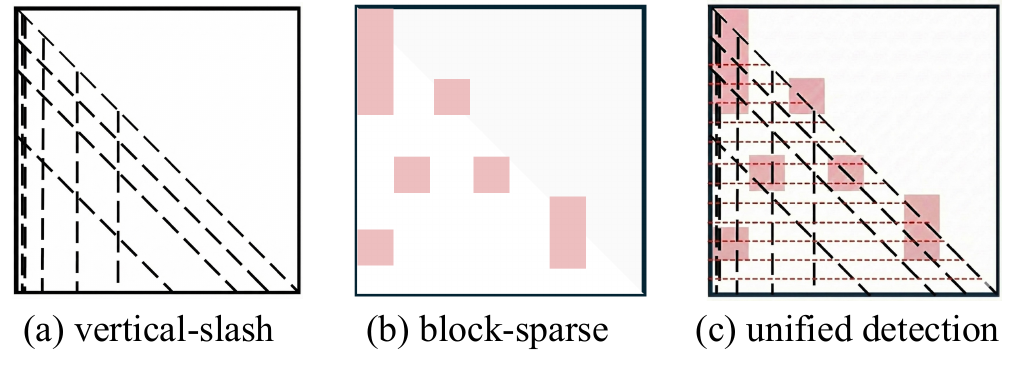}
    \caption{Illustration of pattern discovery. Red dashed lines represent the uniformly distributed queries.}
    \label{fig:attn_pattern}
\end{figure}

\section{Method}
\subsection{Instantaneous Pattern Discovery}
\label{sec:discovery}
Drawing from prior research, we identify three primary sparsity patterns inherent in LLMs: vertical, slash, and block-based sparsity~\cite{minference,flexprefill,xattention}. Most existing methodologies rely on search-based strategies to determine the specific pattern for each attention head, which often introduces additional computational profiling overhead~\cite{minference,flexprefill}. Conversely, non-search-based alternatives typically require more intensive computation and the maintenance of dense attention score matrices, leading to substantial computational and memory access costs~\cite{xattention}.

To facilitate the rapid identification of sparsity patterns within attention maps, we perform a qualitative analysis of various sparsity structures. As illustrated in Fig.~\ref{fig:attn_pattern}, a skeletal set of uniformly distributed queries is sufficient to concurrently resolve vertical, slash, and block-sparse patterns. We attribute the effectiveness of this approach to the following three structural properties:
\begin{itemize}
    \item Vertical Patterns (Column-wise Invariance): These structures represent global key salience, where specific "anchor" tokens attract significant attention regardless of the query's position. Because these features are column-wise invariant, a sparse probing grid acts as a sufficient observer to pinpoint these high-energy columns.
    \item Slash Patterns (Translational Symmetry): Diagonal motifs typically emerge from local syntactic dependencies and relative positional biases. Due to their translational symmetry across the sequence, uniformly distributed probes can effectively sample these local "slices" and resolve the existence of continuous diagonal structures.
    \item Block-sparse Patterns (Spatial Contiguity): These patterns are characterized by localized energy clusters. Leveraging the spatial contiguity of these regions, a uniform sampling strategy ensures a high probability of intersection with these dense clusters, allowing the importance of an entire block to be robustly inferred from a sparse "hit."
\end{itemize}
Based on these insights, we directly employ the full set of queries to compute attention scores against all keys. To ensure maximum accuracy, we opt for this exhaustive approach to precisely identify the salient blocks.

However, Direct query-key interaction over long sequences is computationally prohibitive. To optimize, we employ average-pooled keys $\bar{k} = \frac{1}{n}\sum_{k_i \in \mathcal{B}} k_i$ as block-level proxies. This approach is predicated on the semantic locality and local coherence inherent in LLM embedding spaces, where tokens within a localized block exhibit high feature similarity and redundant attention patterns~\cite{zhang2023h2o,beltagy2020longformer,optimizingmixtureblockattention}. Mathematically, let a block $\mathcal{B}$ consist of $n$ keys $\{k_1, \dots, k_n\}$ with logits $x_i = q \cdot k_i$. We define the probing score $\Psi_{\text{pool}}$ and the true contribution $\Psi_{\text{sum}}$ as follows:
$$\begin{aligned}
\Psi_{\text{pool}} &= \exp(q \cdot \bar{k}) = \exp\left( \frac{1}{n} \sum_{i=1}^n x_i \right) \\
&= \left( \prod_{i=1}^n e^{x_i} \right)^{1/n}= \text{GM}(e^{x_1}, \dots, e^{x_n})
\end{aligned}$$
$$\Psi_{\text{sum}} = \sum_{i=1}^n e^{x_i} = n \cdot \text{AM}(e^{x_1}, \dots, e^{x_n})$$
where $\text{GM}(\cdot)$ and $\text{AM}(\cdot)$ denote the geometric and arithmetic means, respectively. By the AM-GM Inequality:
$$\frac{1}{n} \Psi_{\text{sum}} \geq \Psi_{\text{pool}}$$
While $\Psi_{\text{pool}}$ serves as a lower bound, the rank-order invariance across blocks is maintained due to the low intra-block variance ($\sigma^2 \to 0$) characteristic of attention distributions. In this regime, the GM acts as a strictly monotonic proxy for the AM, ensuring that the relative ordering of blocks remains unchanged with negligible rank distortion. Upon computing the attention scores for individual queries, we perform a secondary averaging operation across all queries within each block to derive the aggregate block-level significance.

\begin{figure}[t]
    \centering
    \includegraphics[width=0.8\linewidth]{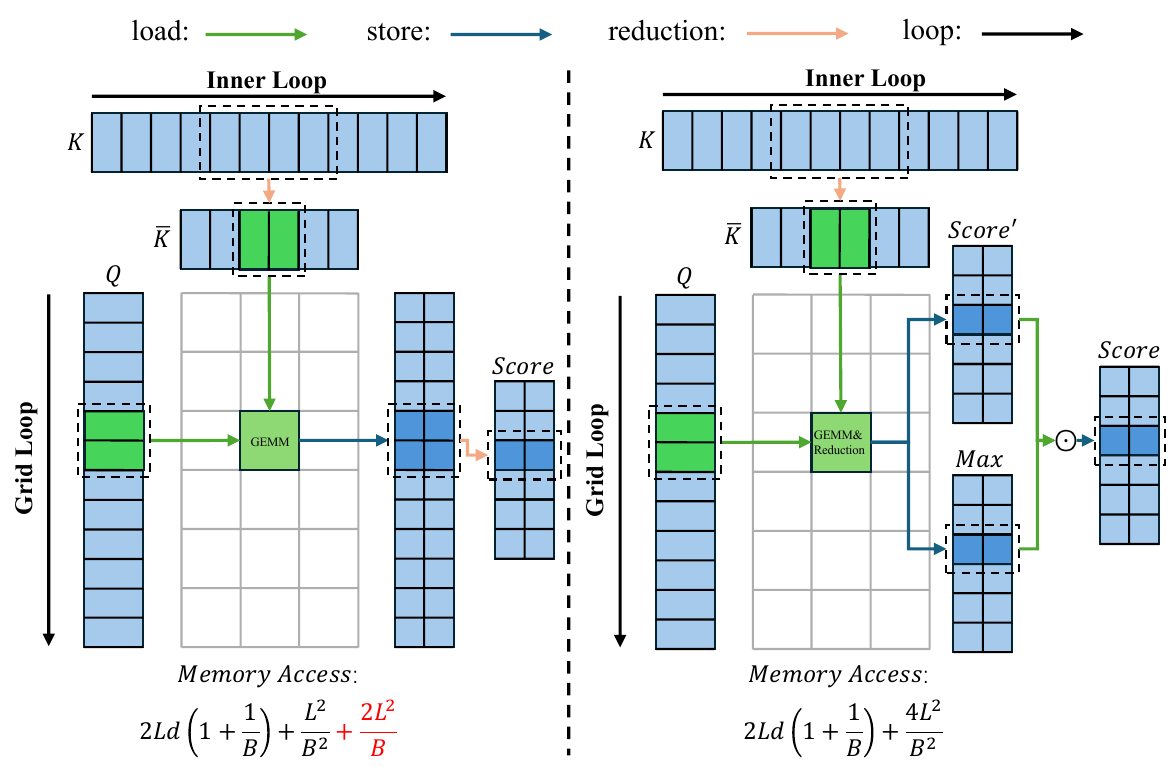}
    \caption{Comparison of block scoring methodologies.(Left) Standard kernel with explicit attention score materialization.(Right) Our optimized kernel utilizing block-wise attention approximation. Compared to the standard implementation, our proposed block-level attention score approximation significantly reduces memory access overhead by bypassing the $O(L^2/B)$ intermediate memory traffic. Here, $L$ denotes the sequence length, $B$ represents the block size, and $d$ indicates the token dimension.}
    \label{fig:probe_kernel}
\end{figure}

\subsection{Block Approximation for Kernel Optimization}
\label{sec:bako}
However, above approach still entails the storage and computation of a massive $L \times (L/B)$ intermediate matrix when calculating intra-block attention averages, where $B$ is block size. To address this, we once again leverage the semantic similarity of tokens within the same block~\cite{zhang2023h2o,beltagy2020longformer,optimizingmixtureblockattention}. Given that tokens in a localized block exhibit significant coherence, their resulting attention distributions are highly redundant. Based on this consideration, we depart from the computationally expensive strict attention calculation in favor of a block-wise attention score approximation. Specifically, we implement a Fused 2D-Reduction kernel that transitions the computation from an explicit "compute-then-pool" sequence to a single-pass fused kernel, significantly reducing global memory traffic. Fig.~\ref{fig:probe_kernel} compares the two methods; we detail our approach below.

\paragraph{Fused Block-level Attention Approximation.}
To avoid the prohibitive cost of strict token-level calculation, we depart from exact attention in favor of a block-wise approximation. By setting the query tile size equal to the block size ($B$), the kernel computes interactions between query tiles and pooled key blocks. We utilize a 1D reduction across the query dimension to compress fine-grained scores into a single approximated scalar per block-pair within the GPU's SRAM:
\begin{itemize}
\item \textbf{Tiled Interaction:} For each query tile, the kernel loads pooled key blocks (where each key in the block is represented by a single vector $k_J$) and computes the interaction $Qk_J^\top$ as a representative proxy.
\item \textbf{Stable Online Reduction:} To maintain numerical stability while approximating Softmax behavior, we perform a max-reduction and exponentiation along the query axis:
\begin{equation}
m_{I,J} = \max_{q_i \in \text{Tile}I} (q_i \cdot k_J)
\end{equation}
\begin{equation}
\mathcal{S}_{I,J} = \sum_{q_i \in \text{Tile}I} \exp(q_i \cdot k_J - m_{I,J})
\end{equation}
\end{itemize}
The kernel executes a sum across the query tile dimension to produce the approximated energy score $\mathcal{S}_{I,J}$ and the local maximum $m_{I,J}$ for each block pair $(I, J)$.

\paragraph{Consistency-preserved Global Normalization.} 
While the block-wise approximation $\mathcal{S}_{I,J}$ is stabilized using local maximums, these scores must be aligned to ensure global comparability across the sequence. We perform a global normalization pass to transform these approximated values into a unified importance map:
\begin{itemize}
    \item \textbf{Maximum Rescaling:} We identify the global maximum for each query tile across all blocks, $\mathcal{M}_I = \max_J (m_{I,J})$, and rescale the local approximated scores to a common denominator:
    \begin{equation}
        \mathcal{S}'_{I,J} = \mathcal{S}_{I,J} \times \exp(m_{I,J} - \mathcal{M}_I)
    \end{equation}
    \item \textbf{Probability Mapping:} The final block importance score is derived by normalizing against the total approximated energy of the row:
    \begin{equation}
        \text{Score}_{I,J} = \frac{\mathcal{S}'_{I,J}}{\sum_K \mathcal{S}'_{I,K} + \epsilon}
    \end{equation}
\end{itemize}
This fused process ensures that the resulting block-level map accurately reflects the relative attention distribution while reducing the memory footprint from $O(L \cdot L/B)$ down to $O((L/B)^2)$. By discarding strict calculation for this high-fidelity approximation, we enable near-instantaneous pattern discovery even for ultra-long sequences.

\subsection{Comparison with Previous Methods.}
\label{sec:cpm}
Compared to previous pattern discovery methods, our proposed method is both faster and more accurate. 
Based on Llama-3.1-8B-Instruct, we compared three different pattern discovery methods: 1) applying mean pooling to both query and key blocks followed by calculating inter-block attention scores~\cite{flexprefill,minference}; 2) using the original method described in Sec.~\ref{sec:discovery} without any block approximation described in Sec.~\ref{sec:bako}~\cite{optimizingmixtureblockattention,mobamixtureblockattention}; and 3) our proposed method based on block approximation.
\begin{table}[ht]
    \centering
    \setlength{\tabcolsep}{2.5mm}
    \scalebox{1.0}{
    \begin{tabular}{c|ccc|ccc}
        \toprule[1pt]
        Method & 4K & 16K & 64K & 4K & 16K & 64K  \\
        \midrule[0.5pt]
        1)& 63.12 & 47.23 & 24.32 & 0.20ms & 0.22ms & 0.63ms\\
        2)& 87.69 & 83.12 & 73.26 & 0.68ms & 2.48ms & 18.26ms \\
        3)& 93.12 & 87.13 & 76.08 & 0.22ms & 0.28ms & 2.21ms\\
        \bottomrule[1pt]
    \end{tabular}}
    \caption{Comparison of different methods for pattern discovery.}
    \label{tab:pattern_discovery}
\end{table}

The results are presented in Tab.~\ref{tab:pattern_discovery}. We conducted evaluations on the RULER~\cite{hsieh2024ruler} benchmark; to ensure a fair comparison among the different pattern discovery methods, we consistently selected the top-8 highest-scoring blocks. Method 1) suffers from significant performance degradation due to its overly aggressive estimation strategy. In contrast, Method 2) incurs substantial time overhead because of excessive memory access costs. Our proposed Method 3) achieves the optimal balance between efficiency and effectiveness.

\subsection{Max-based Dynamic Thresholding}
Previous methods rely on Top-$k$ or Top-$p$ heuristics to select salient tokens or blocks~\cite{xattention,flexprefill,minference,mobamixtureblockattention,optimizingmixtureblockattention}. However, these approaches necessitate relatively expensive sorting and cumulative computations, which is sub-optimal in terms of efficiency.
\begin{figure}[t]
    \centering
    \includegraphics[width=0.74\linewidth]{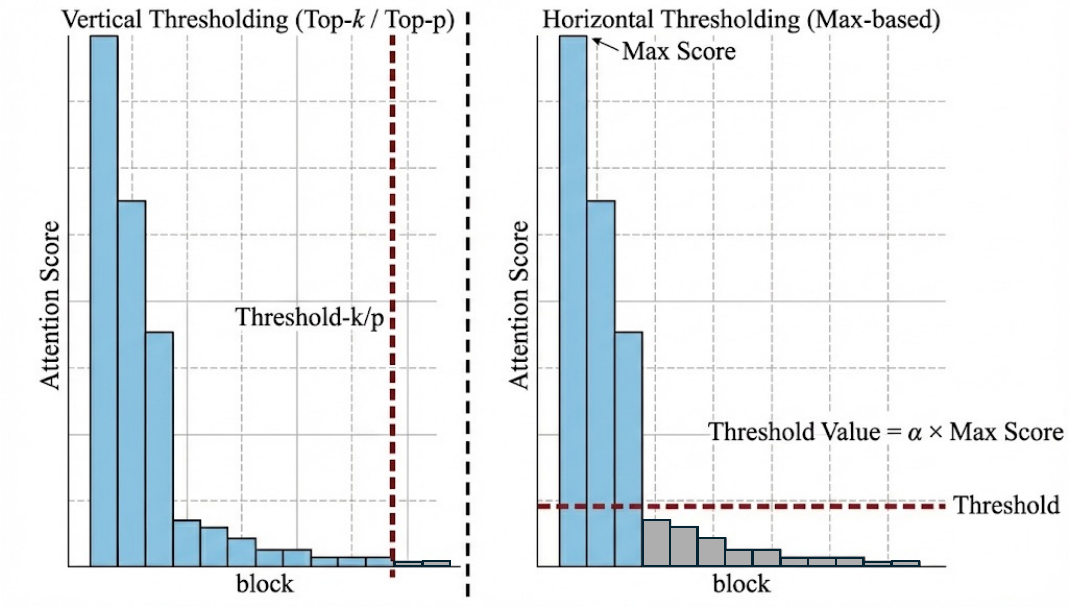}
    \caption{Comparison of different thresholding methodologies. (Left) Top-$k$ and Top-$p$ selection strategies. (Right) Our proposed Max-based thresholding. Top-$k$ and Top-$p$ methods are often susceptible to long-tail distributions, as they may include numerous low-significance blocks simply to satisfy the fixed $k$ or $p$ constraints. In contrast, our dynamic thresholding effectively mitigates the impact of the long tail, achieving higher sparsity through a more precise selection of salient blocks.}
    \label{fig:thresh}
\end{figure}

Furthermore, Top-$p$ and Top-$k$ mechanisms suffer from insufficient sparsity due to their high susceptibility to long-tail distributions. As illustrated in Fig.~\ref{fig:thresh}, when the attention score distribution follows a long tail—which is prevalent in most scenarios—these methods often necessitate the inclusion of numerous insignificant blocks merely to satisfy the fixed $k$ or $p$ constraints, thereby failing to achieve optimal pruning efficiency.

\begin{table*}[t]
    \centering
    \setlength{\tabcolsep}{1.0mm}
    \scalebox{0.9}{
    \begin{tabular}{c|c|cccccccc}
        \toprule[1pt]
        Density& Method & 4K & 8K & 16K & 32K & 64K & 128K & 256K & 512K \\
        \midrule[0.5pt]
        \multirow{2}{*}{60\%}&\cite{guo2024blocksparse} & 1.27ms & 4.16ms & 15.29ms & 59.39ms & 235.67ms & 940.09ms & 3751.57ms & 14990.36ms\\
        &Ours & 0.72ms & 2.80ms & 10.81ms & 43.01ms & 172.05ms & 689.80ms & 2757.41ms & 11256.11ms\\
        \midrule[0.5pt]
        \multirow{2}{*}{6\%}&\cite{guo2024blocksparse} & 0.43ms & 0.55ms & 1.61ms & 6.14ms & 24.48ms & 94.68ms & 383.46ms & 1513.66ms \\
        &Ours & 0.14ms & 0.23ms & 0.98ms & 4.20ms & 17.31ms & 68.48ms & 278.69ms & 1109.44ms\\
        \bottomrule[1pt]
    \end{tabular}}
    \caption{Latency comparison for block-sparse attention implementations.}
    \label{tab:block_sparse_attention}
\end{table*}
\begin{table*}[t]
    \centering
    \setlength{\tabcolsep}{0.8mm}
    \scalebox{0.77}{
    \begin{tabular}{c|ccccccccccc}
    \toprule[1pt]
        Method & En.Sum	& En.QA	& En.MC	& En.Dia & Zh.QA & Code.Debug & Math.Find & Retr.PassKey & Retr.Number & Retr.KV & Avg \\
        \midrule[0.5pt]
        \multicolumn{12}{c}{Llama-3.1-8B-Instruct}\\
        \midrule[0.5pt]
        Full & 32.04 & 25.93 & 69.00 & 21.00 & 31.75 & 18.02 & 25.14 & 99.32 & 99.66 & 62.00 & 48.39 \\
        MInference & 32.04 & 21.94 & 64.63 & 14.50 & 31.75 & 5.33 & 27.43 & 56.61 & 78.31 & 14.00 & 34.65 \\ 
        FlexPrefill & 30.42 & 24.82 & 68.41 & 15.50 & 32.46 & 16.75 & 31.14 & 95.64 & 99.83 & 44.00 & 45.90 \\
        XAttention & 30.10 & 24.79 & 68.56 & 15.50 & 32.28 & 13.71 & 27.14 & 92.54 & 93.90 & 39.00 & 43.75\\
        FlashMoBA & 14.56 & 12.82 & 26.20 & 5.00 & 14.29 & 2.03 & 11.14 & 44.24 & 43.05 & 0.00 & 17.33 \\
        FlashPrefill & 32.04 & 25.07 & 67.69 & 16.50 & 31.22 & 16.75 & 24.86 & 96.10 & 97.97 & 55.00 & 46.32 \\
        \midrule[0.5pt]
        \multicolumn{12}{c}{Qwen2.5-7B-Instruct} \\
        \midrule[0.5pt]
        Full & 17.48 & 5.13 & 44.54 & 15.50 & 8.99 & 18.53 & 34.00 & 0.34 & 94.24 & 0.00 & 23.87 \\
        MInference & 16.50 & 3.70 & 39.74 & 12.50 & 7.94 & 16.24 & 24.29 & 6.78 & 91.36 & 0.00 & 21.90 \\
        FlexPrefill & 18.45 & 4.84 & 48.47 & 9.50 & 7.94 & 17.51 & 29.71 & 4.41 & 95.59 & 0.00 & 23.64\\
        XAttention & 17.48 & 3.99 & 49.78 & 13.50 & 6.88 & 18.78 & 30.00 & 1.36 & 93.73 & 0.00 & 23.55 \\
        FlashMoBA & 2.91 & 1.14 & 11.79 & 6.00 & 2.12 & 7.11 & 13.14 & 0.00 & 41.19 & 0.00 & 8.54\\
        FlashPrefill & 20.39 & 4.84 & 47.16 & 10.50 & 8.99 & 13.20 & 37.43 & 11.19 & 95.59 & 0.00 & 24.93 \\
        \midrule[0.5pt]
        \multicolumn{12}{c}{Qwen3-30B-A3B-Instruct-2507} \\
        \midrule[0.5pt]
        Full & 33.01 & 27.35 & 65.94 & 32.50 & 10.05 & 49.24 & 34.86 & 21.19 & 100.00 & 4.20 & 37.83 \\
        MInference & 25.24 & 22.79 & 59.83 & 17.50 & 7.41 & 33.50 & 30.29 & 22.37 & 92.37 & 6.80 & 31.81 \\ 
        FlexPrefill & 31.07 & 22.22 & 61.57 & 22.50 & 8.47 & 37.31 & 33.14 & 27.46 & 98.14 & 5.40 & 34.73 \\ 
        XAttention & 25.24 & 24.79 & 64.19 & 21.00 & 5.29 & 32.23 & 34.00 & 29.32 & 96.10 & 7.20 & 33.94 \\
        FlashMoBA & 6.80 & 8.26 & 27.95 & 12.00 & 2.65 & 7.61 & 13.14 & 19.15 & 44.24 & 0.00 & 14.18 \\
        FlashPrefill & 33.01 & 23.36 & 63.76 & 19.50 & 9.52 & 34.77 & 36.57 & 33.22 & 100.00 & 8.60 & 36.23 \\
        \bottomrule[1pt]
    \end{tabular}}
    \caption{Performance comparison of different methods on various models and tasks on InfiniteBench.}
    \label{tab:infinitebench}
\end{table*}

To circumvent these limitations, we introduce a Max-based Dynamic Thresholding mechanism. Specifically, for the $I$-th query block, we identify the peak attention score across all candidate key blocks and derive the pruning threshold directly from this maximum value:
$$\text{thresh}_I = \alpha \cdot \max_{J \le I}(\text{Score}_{I, J})$$
where $\alpha$ is a tunable scaling factor. For query block $I$, any key block $J$ whose score falls below $\text{thresh}_I$ is discarded from the computation. Crucially, this approach only requires a single-pass max-reduction instead of computationally expensive global sorting, significantly enhancing execution efficiency. Furthermore, as illustrated in Fig.~\ref{fig:thresh}, this dynamic thresholding effectively mitigates the interference of long-tail distributions, enabling superior sparsity compared to traditional heuristics by focusing exclusively on truly salient blocks.

\subsection{Optimized Block Sparse Attention Kernel}
After obtaining the sparsity pattern between the query and key blocks, FlashPrefill performs standard block-sparse attention on these blocks. In previous works~\cite{xattention,wang2025proxyattn}, this computation was predominantly implemented based on the \textit{Block-Sparse-Attention} codebase~\cite{guo2024blocksparse}. 

However, we observed that the implementation in the \textit{Block-Sparse-Attention} repository fails to fully capitalize on the potential efficiency of sparse execution. The repository employs a logical skipping strategy for masked blocks: its inner loop indiscriminately iterates through the entire linear range of key blocks, using conditional branches to bypass GEMM operations for masked regions. Even when a block is masked and its matrix multiplication is bypassed, the thread execution flow must still fetch, decode, and execute the loop control logic, pointer arithmetic, and synchronization primitives within that iteration. This redundant instruction stream overhead leads to significant pipeline under-utilization, effectively bottlenecking the kernel's overall throughput as the sequence length increases.

To further push the performance frontiers of block-sparse attention, we have optimized its execution model. Specifically, we depart from the logical skipping strategy, which suffers from instruction stream overhead, and implement an index-driven physical jumping mechanism. By directly redirecting memory pointers to salient block coordinates, our approach eliminates redundant control-flow processing and synchronization stalls, thereby maximizing hardware throughput and computational intensity in long-sequence scenarios. Tab.~\ref{tab:block_sparse_attention} benchmarks the efficiency of both methods across varying densities and sequence lengths, where our implementation substantially outpaces existing baseline.

\begin{table*}[t]
    \centering
    \setlength{\tabcolsep}{2.0mm}
    \scalebox{0.8}{
    \begin{tabular}{c|ccccccc|cccccc}
         \toprule[1pt]
         Method & 4K & 8K & 16K & 32K & 64K & 128K & Avg & 4K & 8K & 16K & 32K & 64K & 128K \\
         \midrule[0.5pt]
         \multicolumn{14}{c}{Llama-3.1-8B-Instruct} \\
         \midrule
         Full & 96.07 & 93.94 & 93.51 & 90.68 & 86.29 & 74.25 & 89.12 & $1.00\times$ & $1.00\times$ & $1.00\times$ & $1.00\times$ & $1.00\times$ & $1.00\times$ \\
         MInference & 96.34 & 93.79 & 93.48 & 89.76 & 84.32 & 71.28 & 88.16 & $0.11\times$ & $0.16\times$ & $0.45\times$ & $0.83\times$ & $1.34\times$ & $2.45\times$ \\
         FlexPrefill & 95.63 & 93.14 & 91.92 & 90.36 & 85.49 & 73.23 & 88.30 & $0.11\times$ & $0.33\times$ & $0.98\times$ & $2.21\times$ & $4.16\times$ & $5.18\times$\\
         XAttention & 95.01 & 92.16 & 91.91 & 90.06 & 85.17 & 71.68 & 87.67 & $0.77\times$ & $1.27\times$ & $1.83\times$ & $2.34\times$ & $3.19\times$ & $3.48\times$ \\
         FlashMoBA & 93.14 & 83.23 & 79.28 & 72.13 & 60.12 & 31.26 & 69.86 & $0.75\times$ & $1.16\times$ & $1.99\times$ & $3.34\times$ & $4.70\times$ & $5.98\times$ \\
         FlashPrefill & 97.27 & 96.20 & 94.97 & 92.21 & 84.93 & 75.31 & 90.15 & $1.71\times$ & $2.81\times$ & $4.63\times$ & $7.48\times$ & $13.62\times$ & $22.67\times$\\
         \midrule[0.5pt]
         \multicolumn{14}{c}{Qwen2.5-7B-Instruct} \\
         \midrule[0.5pt]
         Full & 95.21 & 93.17 & 92.18 & 90.14 & 71.87 & 25.23 & 77.97 & $1.00\times$ & $1.00\times$ & $1.00\times$ & $1.00\times$ & $1.00\times$ & $1.00\times$ \\
         MInference & 95.61 & 93.21 & 91.62 & 90.16 & 68.22 & 24.38 & 77.20 & $0.11\times$ & $0.16\times$ & $0.45\times$ & $0.86\times$ & $1.39\times$ & $2.88\times$\\
         FlexPrefill & 95.27 & 93.21 & 91.81 & 89.12 & 70.06 & 26.33 & 77.63 & $0.12\times$ & $0.34\times$ & $0.95\times$ & $2.29\times$ & $4.00\times$ & $5.86\times$ \\
         XAttention & 94.52 & 92.14 & 89.72 & 86.21 & 66.65 & 26.36 & 75.93 & $0.84\times$ & $1.56\times$ & $2.32\times$ & $3.04\times$ & $3.64\times$ & $4.34\times$ \\
         FlashMoBA & 92.36 & 83.68 & 76.11 & 68.12 & 49.31 & 20.13 & 64.95 & $0.73\times$ & $1.14\times$ & $1.97\times$ & $3.34\times$ & $4.70\times$ & $5.98\times$ \\
         FlashPrefill & 95.37 & 93.64 & 91.98 & 88.25 & 70.45 & 34.10 & 78.97 & $1.72\times$ & $2.73\times$ & $4.51\times$ & $6.02\times$ & $10.68\times$ & $16.87\times$ \\
         \midrule[0.5pt]
         \multicolumn{14}{c}{Qwen3-30B-A3B-Instruct-2507} \\
         \midrule[0.5pt]
         Full & 95.78 & 95.79 & 95.82 & 94.31 & 90.25 & 87.71 & 93.28 & $1.00\times$ & $1.00\times$ & $1.00\times$ & $1.00\times$ & $1.00\times$ & $1.00\times$ \\
         MInference & 95.53 & 95.12 & 94.87 & 92.62 & 88.63 & 82.16 & 91.48 & $0.11\times$ & $0.16\times$ & $0.45\times$ & $0.85\times$ & $1.39\times$ & $2.73\times$\\
         FlexPrefill & 95.67 & 95.42 & 95.23 & 93.62 & 89.62 & 86.12 & 92.61 & $0.12\times$ & $0.34\times$ & $0.98\times$ & $2.31\times$ & $3.92\times$ & $5.43\times$\\
         XAttention & 94.72 & 93.68 & 92.32 & 90.44 & 88.12 & 83.69 & 90.50 & $0.84\times$ & $1.43\times$ & $2.02\times$ & $2.45\times$ & $2.97\times$ & $3.45\times$ \\
         FlashMoBA & 92.18 & 89.13 & 84.22 & 76.13 & 67.12 & 60.12 & 78.15 & $0.70\times$ & $1.13\times$ & $1.97\times$ & $3.34\times$ & $4.70\times$ & $5.98\times$ \\
         FlashPrefill & 95.76 & 95.89 & 95.03 & 94.06 & 90.11 & 85.20 & 92.68 & $1.71\times$ & $2.79\times$ & $4.48\times$ & $6.94\times$ & $11.45\times$ & $18.67\times$ \\
         \bottomrule[1pt]
    \end{tabular}}
    \caption{Performance vs. efficiency across different models and methods. Evaluation scores (left) and operator speedup relative to full attention (right) are reported.}
    \label{tab:ruler}
\end{table*}

\begin{table}[t]
    \centering
    \setlength{\tabcolsep}{2.5mm}
    \begin{tabular}{c|ccc|c}
    \toprule[1pt]
         Method & Short & Medium & Long & Avg \\
         \midrule[0.5pt]
         \multicolumn{5}{c}{Qwen2.5-VL-7B-Instruct} \\
         \midrule[0.5pt]
         Full & 75.78 & 62.33 & 53.11 & 63.74 \\
         MInference & 71.00 & 61.89 & 51.78 & 61.56 \\
         FlexPrefill & 72.44 & 61.33 & 52.67 & 62.15 \\
         XAttention & 74.11 & 60.30 & 50.13 & 61.51 \\
         FlashPrefill & 74.78 & 61.67 & 53.22 & 63.22 \\
         \midrule[0.5pt]
         \multicolumn{5}{c}{Qwen3-VL-30B-A3B-Instruct} \\
         \midrule[0.5pt]
         Full & 81.11 & 71.44 & 63.78 & 72.11 \\
         MInference & 79.89 & 70.89 & 62.89 & 71.22 \\
         FlexPrefill & 79.33 & 71.00 & 62.67 & 71.01 \\
         XAttention & 79.67 & 70.11 & 62.06 & 70.61 \\
         FlashPrefill & 80.78 & 71.89 & 63.33 & 72.00 \\
    \bottomrule[1pt]
    \end{tabular}
    \caption{Performance comparison of different methods on various models on VideoMME.}
    \label{tab:videomme}
\end{table}

\section{Experiments}
To substantiate the efficacy of FlashPrefill, we conducted a comprehensive suite of experiments. We validated its performance in LLMs on two widely-recognized long-context benchmarks: RULER~\cite{hsieh2024ruler} and InfiniteBench~\cite{InfiniteBench}. Furthermore, we demonstrated its versatile applicability to VLMs through evaluations on the VideoMME~\cite{videomme} benchmark. Additionally, ablation studies were performed to investigate the impact of individual modules on both model effectiveness and execution efficiency. Detailed hyperparameter configurations are provided in the \textcolor{red}{appendix}.

\subsection{Accuracy and Efficiency Results}

\paragraph{Baselines.}For LLMs, we evaluated the performance of FlashPrefill on three distinct models: Llama-3.1-8B-Instruct~\cite{llama3}, Qwen2.5-7B-Instruct~\cite{qwen25technicalreport}, and Qwen3-30B-A3B-Instruct-2507~\cite{qwen3technicalreport}. For comparison, we selected Full Attention~\cite{attention}, MInference~\cite{minference}, FlexPrefill ($\gamma=0.9, \tau=0.1$)~\cite{flexprefill}, XAttention ($Stride=16, \tau=0.9$)~\cite{xattention}, and FlashMoBA ($B=128, topk=8$)~\cite{mobamixtureblockattention,optimizingmixtureblockattention} as baselines. All benchmarks and efficiency metrics were measured under identical hyperparameter settings. For VLMs, we selected Qwen2.5-VL-7B-Instruct and Qwen3-VL-30B-A3B-Instruct for evaluation. The hyperparameters for each baseline remain identical to those used in the LLM evaluations. All efficiency metrics were measured on NVIDIA H20 GPUs. FlashAttention 2.8.3 is utilized as the full-attention baseline in our experiments.

\paragraph{InfiniteBench.}Tab.~\ref{tab:infinitebench} presents the results of various models and methods on the InfiniteBench benchmark. It can be observed that FlashPrefill consistently achieves superior performance compared to other baselines across both dense and Mixture-of-Experts (MoE) models.

\begin{table}[t]
    \centering
    \setlength{\tabcolsep}{2.5mm}
    \scalebox{1.00}{
    \begin{tabular}{c|ccccccc}
    \toprule[1pt]
        Method & 4K & 8K & 16K & 32K & 64K & 128K & 256K \\
        \midrule[0.5pt]
        FlexPrefill & 20.4\% & 15.4\% & 14.1\% & 12.3\% & 10.1\% & 8.4\% & 8.4\%\\
        XAttention & 57.4\% & 45.3\% & 37.8\% & 31.1\% & 25.6\% & 21.0\% & 18.5\%\\
        FlashPrefill & 70.4\% & 46.0\% & 29.0\% & 17.6\% & 10.0\% & 5.8\% & 3.5\% \\
    \bottomrule[1pt]
    \end{tabular}}
    \caption{Density of various methods on the Qwen3-30B-A3B-Instruct-2507}
    \label{tab:density}
\end{table}

\paragraph{RULER.}Tab.~\ref{tab:ruler} presents the results of various methods across different models on the RULER benchmark. The left side of the table displays the RULER scores, while the right side reports the operator speedup relative to full attention at different sequence lengths. Compared to prior approaches, FlashPrefill achieves a significant speedup across all tested lengths. Specifically, at a 128K context length, FlashPrefill attains speedups of $\mathbf{22.67\times}$, $\mathbf{16.87\times}$, and $\mathbf{18.67\times}$ on three representative models, respectively, substantially outperforming existing methods.

\paragraph{VideoMME.}Tab.~\ref{tab:videomme} presents the performance of various methods on the Video-MME benchmark. Compared to existing sparse attention approaches, FlashPrefill achieves superior results.

\paragraph{Density.}Tab.~\ref{tab:density} presents the density of various methods on the Qwen3-30B-A3B-Instruct-2507 model. It is observed that FlashPrefill effectively mitigates the impact of the long-tail distribution. As the sequence length increases, the amount of effective information relatively diminishes; consequently, FlashPrefill exhibits a significant reduction in density compared to the other two methods.

\paragraph{End-to-End Time-to-First-Toke Speedup.}To evaluate the acceleration of FlashPrefill during the full LLM prefilling process, we integrated it into the vLLM inference framework and measured the Time-to-First Token (TTFT). The results, summarized in Tab.~\ref{tab:ttft}, demonstrate the end-to-end TTFT speedup across three language models, including both dense and Mixture-of-Experts (MoE) architectures. Compared to Full Attention, FlashPrefill achieves significant performance gains. Notably, on the Qwen3-30B-A3B-Instruct-2507, it delivers a $5.02\times$ speedup at a sequence length of 128K.
\begin{table}[t]
    \centering
    \setlength{\tabcolsep}{2.5mm}
    \scalebox{1.0}{
    \begin{tabular}{c|cccccc}
    \toprule[1pt]
        Method & 4K & 8K & 16K & 32K & 64K & 128K \\
        \midrule[0.5pt]
        \multicolumn{7}{c}{Llama-3.1-8B-Instruct} \\
        \midrule[0.5pt]
        Full & 471ms & 976ms & 2189ms & 5398ms & 14774ms & 45464ms \\
        FlashPrefill & 447ms & 913ms & 1860ms & 3618ms & 7481ms & 15061ms \\
        speedup & $1.05\times$ & $1.07\times$ & $1.18\times$ & $1.49\times$ & $1.97\times$ & $3.02\times$ \\
        \midrule[0.5pt]
        \multicolumn{7}{c}{Qwen2.5-7B-Instruct} \\
        \midrule[0.5pt]
        Full & 423ms & 912ms & 2048ms & 4735ms & 12563ms & 37239ms \\
        FlashPrefill & 406ms & 851ms & 1707ms & 3534ms & 7323ms & 15213ms\\
        speedup & $1.04\times$ & $1.07\times$ & $1.20\times$ & $1.34\times$	& $1.72\times$ & $2.45\times$ \\
        \midrule[0.5pt]
        \multicolumn{7}{c}{Qwen3-30B-A3B-Instruct-2507} \\
        \midrule[0.5pt]
        Full & 267ms & 613ms & 1603ms & 4635ms & 15203ms & 53752ms \\
        FlashPrefill & 257ms & 519ms & 1090ms & 2331ms & 4909ms & 10702ms \\
        speedup & $1.04\times$ & $1.18\times$ & $1.47\times$ & $1.98\times$	& $3.10\times$ & $5.02\times$ \\
    \bottomrule[1pt]
    \end{tabular}}
    \caption{End-to-end TTFT speedup achieved by FlashPrefill.}
    \label{tab:ttft}
\end{table}

\subsection{Ablation Study}
\paragraph{Pattern Discovery \& Thresholding.}The combined process of pattern discovery and thresholding is a prerequisite for sparse attention computation, governed by two primary considerations: 1) the accuracy in identifying critical tokens to prevent performance degradation, and 2) the simplicity and speed of the execution. We discuss the first objective in Sec.~\ref{sec:cpm}, where our proposed block approximation demonstrates superior performance and achieves a robust balance between efficiency and effectiveness compared to structurally similar methods. Regarding the second consideration, we compared the combined execution time of discovery and thresholding across different approaches, with the results summarized in Fig.~\ref{fig:dis_thr}. It is evident that FlashPrefill is significantly faster than other methods in the process of identifying important tokens.

\paragraph{Different Thresholding Approaches.}Previous methods predominantly employ Top-$k$ or Top-$p$ strategies for dynamic token selection; however, these approaches are highly susceptible to long-tail distribution effects as sequence lengths increase. In contrast, our Max-based Dynamic Thresholding eliminates the influence of such long-tail distributions, achieving more thorough sparsity in long-context scenarios. Tab.~\ref{tab:thresholding} presents the performance of Llama-3.1-8B-Instruct on the RULER benchmark using various thresholding methods. The left side of the table reports model scores, while the right side displays the corresponding attention density. It is evident that our proposed method significantly reduces computational density while preserving the vast majority of the model's performance, thereby achieving superior acceleration.
\begin{table}[t]
    \centering
    \setlength{\tabcolsep}{2.5mm}
    \scalebox{1.0}{
    \begin{tabular}{c|ccc|ccc}
    \toprule[1pt]
        Method & 32K & 64K & 128K & 32K & 64K & 128K \\
        \midrule[0.5pt]
        Top-$k$ & 91.08 & 81.67 & 70.22 & 12.5\% & 12.5\% & 12.5\%\\
        Top-$p$ & 92.38 & 82.12 & 72.83 & 17.8\% & 15.7\% & 14.0\%\\
        Ours & 92.21 & 84.93 & 75.31 & 16.0\% & 8.2\% & 4.5\%\\
    \bottomrule[1pt]
    \end{tabular}}
    \caption{Performance scores and attention density across various thresholding strategies.}
    \label{tab:thresholding}
\end{table}
\section{Conclusion}
In this paper, we present FlashPrefill, a novel approach designed to significantly accelerate the long-context prefilling stage of Large Language Models (LLMs). FlashPrefill introduces an instantaneous pattern discovery mechanism, further optimized by a block-approximation-based kernel implementation to minimize memory access overhead. Additionally, we propose Max-based Dynamic Thresholding, which obviates the need for sorting and accumulation operations while mitigating the impact of long-tail distributions, thereby achieving a higher degree of sparsity. Our extensive evaluations across various LLMs, VLMs, and diverse benchmarks demonstrate that FlashPrefill effectively enhances prefill efficiency while maintaining superior performance.

\bibliography{main}

\newpage
\appendix

\section{Hyperparameter Configurations}
In FlashPrefill, the sole hyperparameter requiring adjustment is the $\alpha$ value used to determine the dynamic threshold. We calibrate this threshold by observing the computational density of different models at a sequence length of 4K. Specifically, we regulate $\alpha$ to maintain a computational density of approximately 70\% for 4K sequences, as summarized in Tab.~\ref{tab:density}. We employ a uniform block size of 128 across all models. In addition to the blocks selected via Max-based Dynamic Thresholding, we explicitly retain the attention sinks and a local window, with their sizes set to 256 and 512 tokens, respectively. 
\begin{table}[h]
    \centering
    \begin{tabular}{cccccc}
        \toprule[1pt]
        4K & 8K & 16K & 32K & 64K & 128K \\
        \midrule[0.5pt]
        \multicolumn{6}{c}{Llama-3.1-8B-Instruct($\alpha=0.18$)}\\
        \midrule[0.5pt]
        71.0\% & 45.8\% & 28.0\% & 16.0\% & 8.2\% & 4.5\% \\
        \midrule[0.5pt]
        \multicolumn{6}{c}{Qwen2.5-7B-Instruct($\alpha=0.08$)} \\
        \midrule[0.5pt]
        70.0\% & 46.8\% & 29.2\% & 20.8\% & 10.6\% & 6.6\% \\
        \midrule[0.5pt]
        \multicolumn{6}{c}{Qwen3-30B-A3B-Instruct-2507($\alpha=0.12$)} \\
        \midrule[0.5pt]
        70.4\% & 46.0\% & 29.0\% & 17.6\% & 10.0\% & 5.8\% \\
        \bottomrule[1pt]
    \end{tabular}
    \caption{Hyperparameter configurations and resulting computational density for various models.}
    \label{tab:placeholder}
\end{table}

\section{Detailed Implementation}
To provide a clearer elucidation of our proposed FlashPrefill, we decompose the framework into three distinct stages: (i) Instantaneous Pattern Discovery, (ii) Max-based Dynamic Thresholding and Block Selection, and (iii) Block Sparse Attention. The detailed logic for each stage is formalized as pseudocode in Alg.~\ref{alg:flashprefill_state1}, Alg.~\ref{alg:flashprefill_state2}, and Alg.~\ref{alg:sparse_attention}, respectively.
\begin{algorithm}[h]
  \caption{Instantaneous Pattern Discovery}
  \label{alg:flashprefill_state1}
  \begin{algorithmic}
    \STATE {\bfseries Input:} Query $Q \in \mathbb{R}^{L \times d}$, Key $K \in \mathbb{R}^{L \times d}$, scaling $\tau$, block size $B$
    \STATE {\bfseries Output:} Block-level importance map $\mathbf{P} \in \mathbb{R}^{\frac{L}{B} \times \frac{L}{B}}$
    
    \STATE \COMMENT{// Stage 1: Key Block Pre-pooling}
    \FOR{$J=0$ {\bfseries to} $\frac{L}{B}-1$}
      \STATE $K_{block} = K[J \cdot B : (J+1) \cdot B, : ]$
      \STATE $\bar{k}_J = \text{Mean}(K_{block}, \text{axis}=0)$ \COMMENT{Represent block by a vector}
    \ENDFOR

    \STATE \COMMENT{// Stage 2: Fused Block Approximation}
    \FOR{$I=0$ {\bfseries to} $\frac{L}{B}-1$}
      \STATE $Q_I = Q[I \cdot B : (I+1) \cdot B, : ]$ \COMMENT{Load query tile}
      \FOR{each $\bar{k}_J$ {\bfseries where} $J \le I$}
        \STATE $\mathbf{qk} = (Q_I \cdot \bar{k}_J^\top) \cdot \tau \cdot \log_2 e$ \COMMENT{Result size $B \times 1$}
        \STATE $\mathbf{qk} = \text{ApplyCausalMask}(\mathbf{qk})$
        \STATE $m_{I,J} = \max_{i \in \{0, \dots, B-1\}} (\mathbf{qk}_i)$
        \STATE $\mathcal{S}_{I,J} = \sum_{i=0}^{B-1} 2^{(\mathbf{qk}_i - m_{I,J})}$
      \ENDFOR
    \ENDFOR

    \STATE \COMMENT{// Stage 3: Consistency-preserved Global Normalization}
    \FOR{$I=0$ {\bfseries to} $\frac{L}{B}-1$}
      \STATE $\mathcal{M}_I = \max_{J \le I} (m_{I,J})$ \COMMENT{Global max for row $I$}
      \FOR{$J=0$ {\bfseries to} $I$}
        \STATE $\mathcal{S}'_{I,J} = \mathcal{S}_{I,J} \times 2^{(m_{I,J} - \mathcal{M}_I)}$
      \ENDFOR
      \STATE $\mathbf{P}_{I,J} = \mathcal{S}'_{I,J} / (\sum_{K=0}^{I} \mathcal{S}'_{I,K} + \epsilon)$
    \ENDFOR
    \STATE {\bfseries return} $\mathbf{P}$
  \end{algorithmic}
\end{algorithm}

\begin{algorithm}[h]
  \caption{Max-based Dynamic Thresholding and Block Selection}
  \label{alg:flashprefill_state2}
  \begin{algorithmic}
    \STATE {\bfseries Input:} Block scores $\mathbf{S} \in \mathbb{R}^{Z \times M \times N \times H}$, sink size $S$, window size $W$, threshold factor $\alpha$
    \STATE {\bfseries Output:} Compact indices $\mathbf{I}$, Active counts $\mathbf{C}$

    \STATE \COMMENT{// Step 1: Generate Dynamic and Structural Masks}
    \FOR{each batch $z$, query block $i$, and head $h$}
      \STATE $max\_val = \max_{j \in \{0, \dots, N-1\}} \mathbf{S}_{z,i,j,h}$
      \FOR{each key block $j \in \{0, \dots, N-1\}$}
        \STATE $mask\_score = (\mathbf{S}_{z,i,j,h} \ge max\_val \cdot \alpha)$
        \STATE $mask\_sink = (j < S)$
        \STATE $mask\_window = (0 \le i - j < W)$
        \STATE $mask\_causal = (i \ge j)$
        \STATE \COMMENT{// Union of active blocks limited by causality}
        \STATE $is\_active_{z,i,j,h} = (mask\_score \lor mask\_sink \lor mask\_window) \land mask\_causal$
      \ENDFOR
    \ENDFOR

    \STATE \COMMENT{// Step 2: Index Compression}
    \FOR{each batch $z$, query $i$, and head $h$}
      \STATE $\mathbf{C}_{z,i,h} = \sum_{j=0}^{N-1} is\_active_{z,i,j,h}$ \COMMENT{Sum of active blocks}
      \FOR{each key block $j$}
        \IF{$is\_active_{z,i,j,h}$ is $True$}
          \STATE $idx\_to\_sort_{z,i,j,h} = j$
        \ELSE
          \STATE $idx\_to\_sort_{z,i,j,h} = N$ \COMMENT{Fill inactive with $N$}
        \ENDIF
      \ENDFOR
      \STATE $\mathbf{I}[z,i,:,h] = \text{Sort}(idx\_to\_sort_{z,i,:,h})$ \COMMENT{Stable sort to compact indices}
    \ENDFOR
    \STATE {\bfseries return} $\mathbf{I}, \mathbf{C}$
  \end{algorithmic}
\end{algorithm}

\begin{algorithm}[h]
  \caption{Block Sparse Attention Kernel}
  \label{alg:sparse_attention}
  \begin{algorithmic}
    \STATE {\bfseries Input:} Query $Q$, Key $K$, Value $V$, Block indices $\mathbf{Idx}$, Valid counts $\mathbf{C}$, scale $\tau$
    \STATE {\bfseries Output:} Output $O$, Log-sum-exp $L$
    
    \FOR{each query tile $I \in \{0, \dots, \frac{L}{Q_{TILE}}-1\}$}
      \STATE \COMMENT{// Initialization for Online Softmax}
      \STATE $m_i = -\infty, \quad \ell_i = 0, \quad acc = 0$
      \STATE $Q_I = \text{Load}(Q, \text{tile } I)$
      \STATE $N_{active} = \mathbf{C}[I]$ \COMMENT{Get count of salient blocks}
      
      \STATE \COMMENT{// Iterate only through identified salient blocks}
      \FOR{$idx = 0$ {\bfseries to} $N_{active}-1$}
        \STATE $B_{id} = \mathbf{Idx}[I, idx]$ \COMMENT{Target block index}
        \STATE $is\_diag = (B_{id} == I )$ \COMMENT{Check for diagonal/causal block}
        
        \STATE \COMMENT{// Process key/value within the salient block}
        \FOR{each key tile $J$ within block $B_{id}$}
          \STATE $K_J, V_J = \text{Load}(K, V, \text{tile } J)$
          \STATE $\mathbf{qk} = (Q_I \cdot K_J^\top) \cdot \tau \cdot \log_2 e$
          
          \IF{$is\_diag$ is $True$}
            \STATE $\mathbf{qk} = \text{ApplyCausalMask}(\mathbf{qk})$
          \ENDIF
          
          \STATE \COMMENT{// Online Softmax Update}
          \STATE $m_{new} = \max(m_i, \max(\mathbf{qk}, \text{axis}=1))$
          \STATE $P = 2^{(\mathbf{qk} - m_{new})}$
          \STATE $\ell_{new} = \ell_i \cdot 2^{(m_i - m_{new})} + \sum(P, \text{axis}=1)$
          
          \STATE $acc = acc \cdot 2^{(m_i - m_{new})} + P \cdot V_J$
          \STATE $m_i = m_{new}, \quad \ell_i = \ell_{new}$
        \ENDFOR
      \ENDFOR
      
      \STATE \COMMENT{// Final Normalization and Storage}
      \STATE $O[I] = acc / \ell_i$
      \STATE $L[I] = m_i + \log_2(\ell_i)$
    \ENDFOR
    \STATE {\bfseries return} $O, L$
  \end{algorithmic}
\end{algorithm}

\end{CJK*}
\end{document}